\documentclass[12pt]{article}

\usepackage{latexsym, amssymb, amsbsy, epsfig, enumerate}
\usepackage{amsmath, amsopn, theorem}
\usepackage{subfigure} 
\usepackage{color}

\usepackage[round]{natbib}

\usepackage[nofiglist, notablist]{endfloat}

{ \theorembodyfont{\rmfamily}

}

\def\m#1{\mbox{#1}}

\def\c1{{\m{1}\!\m{I}}}
\def\calR{{\mathbb R}}

\def\bb{{\bf b}}

\def\bw{{\bf w}}
\def\bx{{\bf x}}

\def\bmu{\boldsymbol{\mu}}

\def\sumin{\sum_{i=1}^n}

\def\sumipn{\sum_{i'=1}^n}
\def\sumjp{\sum_{j=1}^p}

\begin{document}

\title{A robust and sparse K-means clustering algorithm}
\author{Yumi Kondo \and Matias Salibian-Barrera
\and Ruben Zamar \thanks{Department of Statistics, The University of British Columbia, Vancouver, Canada}}

\date{\today}

\maketitle

\noindent Keywords: K-means, robust clustering, sparse clustering, trimmed K-means.

\abstract{
In many situations where 
the interest lies in identifying clusters
one might expect that not all available variables
carry information about these groups.
Furthermore, data quality (e.g. outliers or missing
entries) might 
present
a serious and sometimes hard-to-assess problem for  
large and complex datasets.
In this paper 
we show that a small proportion
of atypical observations might have serious
adverse effects on the solutions found by the 
sparse clustering algorithm of \citet{Dan}. 
We propose a robustification of their
sparse K-means algorithm 
based on the trimmed K-means algorithm of \citet{trimmed}.
Our proposal is also able to 
handle datasets with missing values. 
We illustrate the use of our method on 
microarray data for cancer patients where we are
able to identify strong biological clusters with a much
reduced number of genes. 
Our simulation studies show that,
when there are outliers in the data,
our robust sparse K-means algorithm performs
better than other competing methods both in terms of the selection
of features
and also the identified clusters. 
This robust sparse K-means algorithm is 
implemented in the \verb=R= package \verb=RSKC= which is 
publicly available from the \verb=CRAN= repository.
}

\section{Introduction}

K-means is a broadly used clustering method first introduced by \citet{hugo}
and further popularized by \citet{km}. K-means popularity
derives in part from its conceptual simplicity (it optimizes a very natural
objective function) and widespread implementation in statistical packages.

Unfortunately, it is easy to construct simple examples 
where K-means performs rather poorly in the
presence of a large number of \emph{noise variables}, i.e., variables that
do not change from cluster to cluster. These type of datsets are commonplace
in modern applications. 
Furthermore, in some applications it is of interest to identify
not only possible clusters in the data, but also a relatively
small number of variables that sufficiently determine that structure. 

\begin{figure}[tp]
\begin{center}
\subfigure[True cluster labels]{ 
\includegraphics[height=2.5in, angle=0]{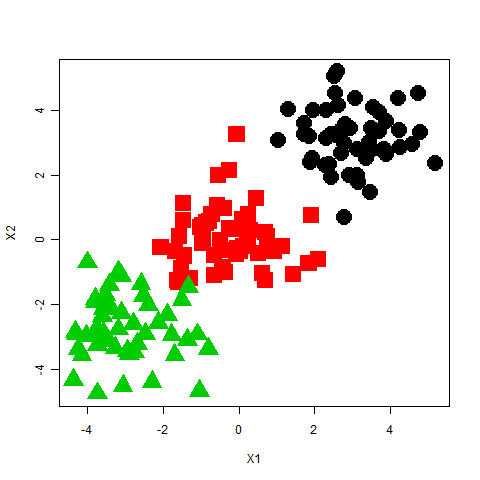}
}
\subfigure[K-Means]{ 
\includegraphics[height=2.5in, angle=0]{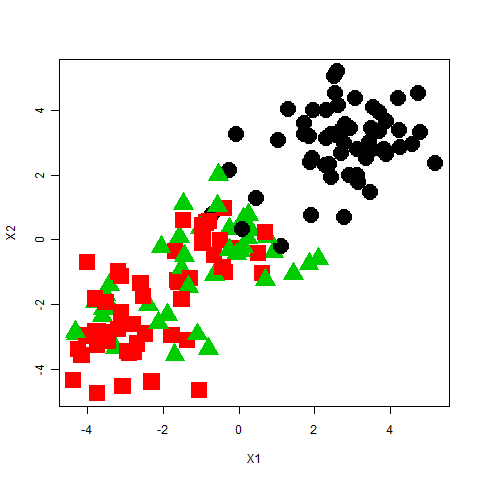}
}
\subfigure[Sparse]{ 
\includegraphics[height=2.5in, angle=0]{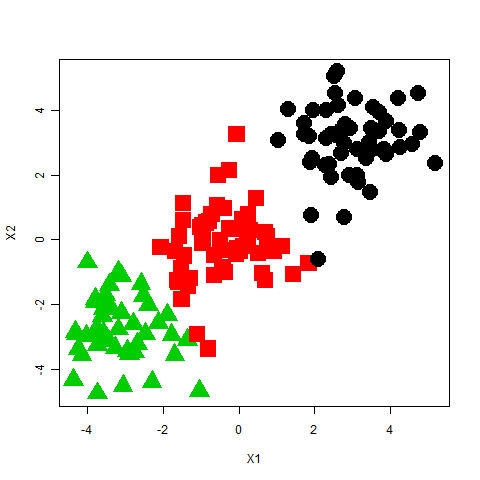}
}
\end{center}
\caption{This is the figure} \label{ex1}
\end{figure}

To address these problems, \citet{Dan} proposed an
alternative to classical K-means - called \emph{sparse K-means (SK-means)} -
which simultaneously finds the clusters and the important \emph{clustering
variables}. 
To illustrate the performance of Sparse K-means 
when there is a large number of noise features
we generate 
$n = 300$ observations with $p=1000$ features. 
The observations follow a multivariate normal distribution 
with covariance matrix equal to the identity and 
centers at $(\mu, \mu, 0, \ldots, 0)$ with $\mu = -3$, 0 and 3 for each
cluster respectively. Only the first 2 features are {\it clustering features}. 
Colors and shapes indicate the true cluster labels.
Figure \ref{ex1} contains the scatter plot of the data with respect to
their clustering features. Panel (a) displays the true cluster labels, while
panels (b) and (c) contain the partition obtained by K-means 
and Sparse K-means
respectively. We see that althought the cluster structure is reasonably 
strong in the first two variables, K-means is unable to detect it, while
Sparse K-means is able to retrieve a very good partition. Moreover, 
the optimal weights chosen by Sparse K-means are positive only for the
clustering features. 

Sparse K-means cleverly exploits the fact that commonly used dissimilarity
measures (e.g. squared Euclidean distance) can be decomposed into $p$ terms,
each of them solely depending on a single variable. More precisely, given
a cluster partition $\mathcal{C=}\left( C_{1},C_{2}, \ldots, C_{K}\right)$,
the associated between-cluster dissimilary measure 
$B\left( \mathcal{C}\right)$ satisfies
\[
B\left( \mathcal{C}\right) =\sum_{j=1}^{p}B_{j}\left( \mathcal{C}\right) \, , 
\]%
where $B_{j}\left( \mathcal{C}\right) $ solely depends on the $j^{th}$
variable.

Given a vector of non-negative weights $\mathbf{w=}\left(
w_{1},w_{2},...,w_{p}\right) ^{\prime }$ --one weight for each variable--
SK-means considers the \emph{weighted between-cluster} dissimilarity measure 
\begin{equation}
B\left( \mathbf{w,}\mathcal{C}\right) =\sum_{j=1}^{p}w_{j}B_{j}\left( 
\mathcal{C}\right) =\mathbf{w}^{\prime }\mathbf{B}\left( \mathcal{C}\right)
\label{SK-means objective function}
\end{equation}%
where 
\[
\mathbf{B}\left( \mathcal{C}\right) \mathbf{=}\left( 
\begin{tabular}{c}
$B_{1}\left( \mathcal{C}\right) $ \\ 
$B_{2}\left( \mathcal{C}\right) $ \\ 
\vdots \\ 
$B_{p}\left( \mathcal{C}\right) $%
\end{tabular}%
\right) . 
\]%
SK-means then searchs for the pair $\left( \mathbf{w}^{\ast }\mathbf{,}%
\mathcal{C}^{\ast }\right) $ that maximizes (\ref{SK-means objective
function}) subject to 
\[
\sum_{i=1}^{p}w_{j}^{2}\leq 1\text{ \ \ and \ \ }\sum_{i=1}^{p}w_{j}\leq
l \, , 
\]%
for some $1 < l < \sqrt{p}$. In other words, SK-means performs a regularized
(LASSO-type) version of K-means.

Unfortunatley, 
as will be illustrated in the next section
and in our simulation studies (see Section \ref{simulations}), Sparse K-means 
may fail to produce
a reasonable partition when the data contain even a very small proportion of
outliers (for example, 1\% of observations having an outlying observation in just one out of several hundred
features). 
To remedy this lack of robustness we propose a robustified
alternative called Robust and Sparse K-means (RSK-means). We will show that
RSK-means is actually robust in the sense that it works well for clean data
and also for data containing outliers. 

The rest of the paper is organized as follows. Section 2 describes
the robust sparse K-means (RSK-means) algorithm and our proposal for
selecting the associated tuning constants. Section 3 illustrates 
our approach on a gene expression data set where RSK-means 
is able to identify biological clusters among breast cancer patients.
Section 4 reports the results of our simulation study and Section 5 
provides concluding remarks. 

\section{Robust and Sparse K-Means}

The K-means clustering algorithm finds $K$ clusters $C_1$, \ldots, $C_K$ that 
minimize the within-clusters sum of squares
\begin{equation} \label{kmeans}
\min_{C_1, \ldots, C_K} \ \left\{ \sum_{k=1}^K \, \frac{1}{n_k} \, \sum_{i, i' \in 
C_k} \, d_{i, i'} \, \right\} \, ,
\end{equation}
where $C_1$, \ldots, $C_K$ are the disjoint sets of cluster indices, 
$n_k$ is the number of observations in the $k$-th cluster, and 
$$
d_{i, i'} = \sum_{j=1}^p d_{i, i', j}
$$ 
is the (additive) dissimilarity measure
between the $i$ and $i'$ observations. 
When our observations are vectors $\bx_1$, $\ldots$, $\bx_n \in \calR^p$
and $d_{i,i'}$ is the squared Euclidean distance between the $i$-th 
and $i'$-th points
have $d_{i, i', j} = ( \bx_{i,j} - \bx_{i', j})^2$, $j=1$, \ldots, $p$
and
$$
\sum_{k=1}^K \, \frac{1}{n_k} \, \sum_{i, i' \in 
C_k} \, \left\| \bx_i - \bx_{i'} \right\|^2 \ = \ 
\sum_{k=1}^K \, \sum_{j \in C_k}
\left\| \bx_j - \mu_k  \right\|^2 \, ,
$$
where $\mu_k$ is the sample mean of the observations in the $k$-th cluster. 
A popular algorithm to find local solutions to \eqref{kmeans} is as follows \citet{lloyd}.
First randomly select K initial ``centers'' $\mu_1$, \ldots, $\mu_K$ and 
iterate the following two steps until convergence:
\begin{enumerate}[(a)]
\item Given cluster centers $\mu_1$, \ldots, $\mu_K$, assign each point to the cluster with the closest
center. 
 \item Given a cluster assignment, update the cluster centers to be the 
sample mean of the observations in each cluster. 
\end{enumerate}
Although this algorithm decreases the objective function at each iteration it 
may be trapped in different local minima. Hence, it is started several times and the best 
solution is returned. 

\citet{trimmed} proposed a modification of this algorithm 
in order to obtain outlier-robust clusters. The main idea is to replace step (b) 
above by
\begin{enumerate}[(a')]
\setcounter{enumi}{1}
\item Given a cluster assignment, trim the $\alpha 100 \%$ observations with 
largest distance to their cluster centers, and update the cluster 
centers to the 
sample mean of the remaining observations in each cluster. 
\end{enumerate}
The tuning parameter $\alpha$ regulates the amount of trimming and is selected by the user.

Since the total sum of squares 
$$
\frac{1}{n} \, \sumin \sumipn d_{i, i'} 
$$
does not depend on the cluster assignments, minimizing the within-cluster sum
of squares is equivalent to maximizing the between-cluster sum of squares
$$
\frac1n \sumin \sumipn d_{i,i'} \ - \ 
\sum_{k=1}^K \, \frac{1}{n_k} \, \sum_{i, i' \in 
C_k} \,  d_{i, i'} \, = \, 
\sumjp \, \left\{ \, \frac1n \sumin \sumipn d_{i,i',j} \ - \ 
\sum_{k=1}^K \, \frac{1}{n_k} \, \sum_{i, i' \in 
C_k} \,  d_{i, i', j} \, \right\} \, .
$$
The SK-means algorithm of \citet{Dan} introduces non-negative weights
$w_j$, $j=1$, \ldots, $p$ for each feature and then solves
\begin{equation} \label{skmeans}
 \max_{C_1, \ldots, C_K, \bw} \ 
\sumjp \, w_j \, \left\{ \, \frac1n \sumin \sumipn d_{i,i',j} \ - \ 
\sum_{k=1}^K \, \frac{1}{n_k} \, \sum_{i, i' \in 
C_k} \,  d_{i, i', j} \, \right\} \, ,  
\end{equation}
subject to $\| \bw \|_2 \le 1$, $\| \bw \|_1 \le l$ and $w_j \ge 0$, $j=1$, \ldots, $p$, 
where $l > 1$ determines the degree of sparcity (in terms of non-zero weights) of the solution. 
%
%
The optimization problem in \eqref{skmeans} can be solved by iterating the following steps:
\begin{enumerate}[(a)]
\item Given weights $\bw$ and cluster centers $\mu_1$, \ldots, $\mu_K$ solve 
$$
 \min_{C_1, \ldots, C_K} \ 
\sum_{k=1}^K \, \sum_{i \in C_k} \sumjp
\, w_j \, (\bx_{i,j} - \mu_{k,j} )^2 \, ,
$$
which is obtained by assigning each point to the cluster with closest center
using weighted Euclidean squared distances. 
 \item Given weights $\bw$ and cluster assignments $C_1$, \ldots, $C_K$, 
update the cluster centers to be the 
weighted sample mean of the observations in each cluster. 
\item Given cluster assignments $C_1$, \ldots, $C_K$ and 
centers $\mu_1$, \ldots, $\mu_K$ solve 
$$
 \max_{\| \bw \|_2 \le 1, \| \bw \|_1 \le l} \ 
\sumjp \, w_j \, B_j( C_1, \ldots, C_K ) \, ,
$$
where $B_j( C_1, \ldots, C_K ) = (1/n) \, \sumin \sumipn d_{i,i',j} \ - \ 
\sum_{k=1}^K \, \frac{1}{n_k} \, \sum_{i, i' \in 
C_k} \,  d_{i, i', j}$. There is 
a closed form expression for the vector of weights that solves this 
optimization problem \citep[see][]{Dan}. 
\end{enumerate}

A naive first attempt to incorporate robustness to this algorithm is to 
use trimmed K-means 
with weighted features, and then 
optimize the weights using the trimmed sample. 
In other words, to replace step (b) above with (b') where 
$B_j(C_1, \ldots, C_K)$, $j=1$, \ldots, $p$ are calculated without
the observations flagged as outliers. 
A potential problem with this approach is 
that if an observation is outlying in a feature that received 
a small weight in steps (a) and (b'), it might not be trimmed. 
In this case, the variable where the outlier is more
evident will receive a very
high weight in step (c) (because this feature will be associated 
with a very large $B_j$). This may in turn cause the weighted K-means 
steps above to form a cluster containing this single point, which 
will then not be downweighted (since its distance to the cluster
centre will be zero). This phenomenom is illustrated in the following example. 

\begin{figure}[tp]
\begin{center}
\subfigure[True cluster labels]{ 
\includegraphics[height=2.5in, angle=0]{example-scatter.png}
}
\subfigure[Naive]{ 
\includegraphics[height=2.5in, angle=0]{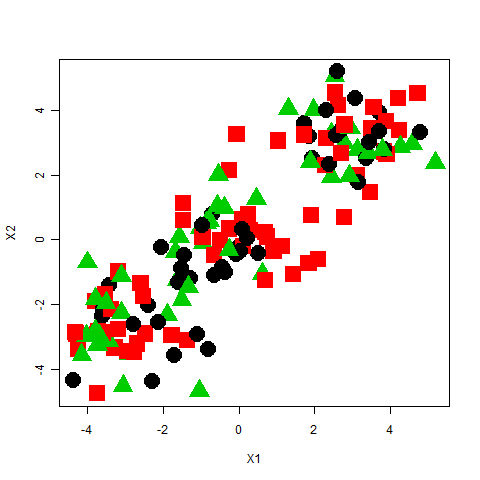}
}
\subfigure[Sparse]{ 
\includegraphics[height=2.5in, angle=0]{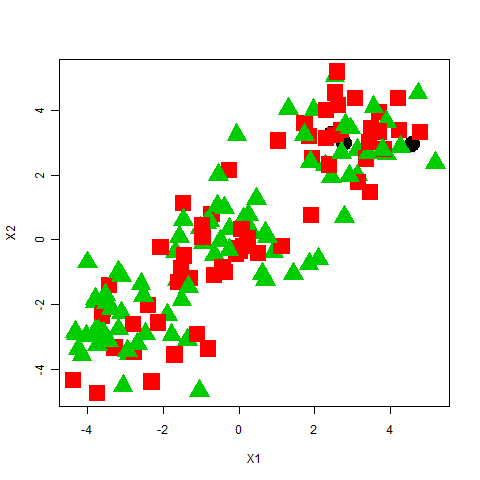}
}
\subfigure[RSK-means]{ 
\includegraphics[height=2.5in, angle=0]{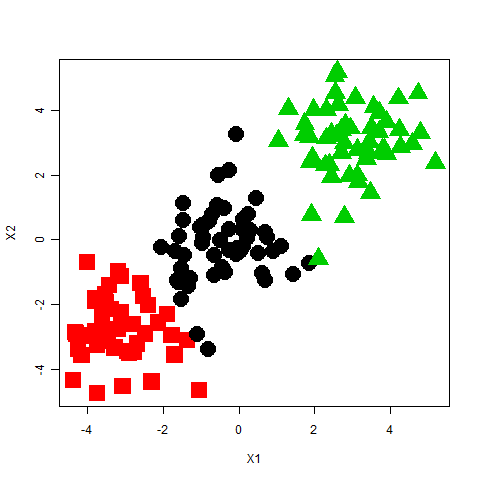}
}
\end{center}
\caption{This is the figure} \label{scatter-1}
\end{figure}

We generated a synthetic data set with $n = 300$ observations and $p=5$
features. The observations follow a multivariate normal distribution 
with covariance matrix equal to the identity and 
centers at $(\mu, \mu, 0, 0, 0)$ with $\mu = -2, 0$ and 2 for each
cluster respectively. Only the first 2 features contain information 
on the clusters. Figure \ref{scatter-1} contains the scatterplot
of these 2 clustering features. Colors and shapes indicate the true cluster labels. 
To illustrate the problem mentioned above, we replaced the
4th and 5th entries of the first 3 observations with
large outliers. The naive robustification described above
returns a dissapointing partition because it fails to
correctly identify the clustering features, placing all
the weight on the noise ones. 
The result is shown in Figure \ref{scatter-1} (b). 
As expected, 
sparse K-means also fails in this case assigning all
the weights to the noise features and forming one small
cluster with the 3 outliers (see Figure \ref{scatter-1} (c)). 
Finally, Figure \ref{scatter-1} (d) shows the partition 
found by the Robust Sparse K-Means algorithm below. 

The key step in our proposal is to
use two sets of trimmed observations which we call
the weighted and un-weighted trimmed sets. 
By doing this, zero weights 
will not necessarily mask outliers in the noise
features. Our algorithm can be described as follows:
\begin{enumerate}[(a)]
\item Perform Trimmed K-means on the weighted 
data set:
\begin{enumerate}[(i)]
\item Given weights $\bw$ and cluster centers $\mu_1$, \ldots, $\mu_K$ solve 
$$
 \min_{C_1, \ldots, C_K} \ 
\sum_{k=1}^K \, \sum_{i \in C_k} \sumjp
\, w_j \, (\bx_{i,j} - \mu_{k,j} )^2 \, ,
$$
which is obtained by assigning each point to the cluster with closest center
using weighted Euclidean squared distances. 

\item Given weights $\bw$ and cluster assignments, 
trim the $\alpha 100 \%$ observations with 
largest distance to their cluster centers, and update the cluster 
centers to the 
sample mean of the remaining observations in each cluster. 

\item Iterate the two steps above until convergence.

\end{enumerate}

\item Let $O_W$ be the subscripts of the $\alpha 100 \%$ 
cases labelled as outliers in the final 
step of the weighted trimmed k-means procedure above. 

\item Using the partition returned by trimmed k-means, 
calculate the unweighted cluster centers $\tilde{\mu}_k$, 
$k=1, \ldots, K$. For each observation $\bx_i$, let $\tilde{d}_i$
be the unweighted distance to its cluster centre, i.e.: 
$\tilde{d}_i = 
\| \bx_i - \tilde{\mu}_k \|^2$ where $i \in C_k$. 
Let $O_E$ be the subscripts of the  
$\alpha 100 \%$ largest distances $\tilde{d}_i$. 

\item Form the set of trimmed points $O = O_W \bigcup O_E$. 

\item Given cluster assignments $C_1$, \ldots, $C_K$, 
centers $\mu_1$, \ldots, $\mu_K$ and trimmed points $O$, 
find a new set of weights $\bw$ by 
solving
$$
 \max_{\| \bw \|_2 \le 1, \| \bw \|_1 \le l} \ 
\sumjp \, w_j \, B_j( C_1, \ldots, C_K, O) \, ,
$$
where $B_j(C_1, \ldots, C_K, O)$, $1 \le j \le p$,  
are calculated without
the observations in $O$. 
\end{enumerate}
We call this algorithm RSK-means, and it is readily
available on-line in the \verb=R= package \verb=RSKC=
from the \verb=CRAN= repository: 
\verb=http://cran.r-project.org=.

The RSK-means algorithm requires the selection of three parameters: the 
$L_1$ bound, the trimming proportion
$\alpha$  and
the number of clusters $K$. Regarding the first two, we 
recommend considering different combinations and comparing
the corresponding results. In 
our experience 
the choice $\alpha = 0.10$
suffices for most applications. The choice of the 
$L_1$ bound determines the 
degree of sparcity and hence can be tuned to achieve
a desired number of selected features. 
In practice, one can also consider several combinations of
these parameters and compare the different results. To
select the number of
clusters $K$ we recommend using the Clest algorithm of \citet{Clest}.
This is the default method implemented in the
\verb=RSKC= package.

\section{Example
- Does this data set have missing values?} 

In this section we consider the problem of identifying clinically interesting
groups and the associated set of discriminating genes among breast cancer
patients using microarray data. We used a data set consisting on 4751 gene
expressions for 78 primary breast tumors.  The dataset was previously analyzed
by \citet{vant} and it is publicly available at:
\texttt{http://www.nature.com/nature/journal/v415/n6871/suppinfo/415530a.html}.
These authors applied a supervised classification technique and found subset of
70 genes that helps to discriminate patients that develop distant metastasis
within 5 years from others. 

To further explore these data we applied both the Sparse and the Robust Sparse
K-means algorithms. We set the $L_1$ bound to 6 in order to obtain
approximately 70 non-zero weights.  As in \citet{cor-disim}, we used a
dissimilarity measure based on Pearson's correlation coefficient between cases.
This can be done easily using the relationship between the squared Euclidean
distance between a pair of standardized rows and their correlation coefficient.
More specifically, let $\tilde{\bx}_i$ denote the standardized $i$-th case,
where we substracted the average of the $i$-th case and divided by the
corresponding standard deviation. Then we have
$$
\| \tilde{\bx}_i - \tilde{\bx_j} \|^2 = 
2 \, p \, \left( 1 - \mbox{corr} \left( \bx_i, \bx_j \right) \right) \, .
$$
Hence, as a dissimilarity measure, we used the
squared Euclidean distance calculated on the standardized data. 
To avoid the potential damaging effect of outliers 
we standardized each observation using 
its median and MAD instead of the sample mean and
standard deviation.

Since we have no information on the number of
potential outliers in these data, we ran Clest
with four trimming proportions: $\alpha=0$, $1/78$, $5/78$ and $10/78$.
Note that using $\alpha=0$ 
corresponds to the original 
sparse $K$-means method. Interestingly, Clest selects $K=2$ for
the four trimming proportions mentioned above. 
%




It is worth noticing that the four algorithms return clusters closely
associated with an important clinical outcome: the estrogen receptor
status of the sample (ER-positive or ER-negative).  There is clinical consensus
that ER positive patients show a better response to treatment 
of metastatic disease \citep[see][]{ERp}. 

To evaluate the strenght of the 
70 identified genes we 
considered the level of agreement between the 
clustering results obtained using these genes and the
clinical outcome. To measure this agreement we use
the classification error rate (CER) (see Section \ref{simulations}
for its definition). 
Since outliers are not considered members of either of the 
two clusters, we exclude them from the CER calculations. 
However, it is important to note that, given a trimming proportion
$\alpha$, the robust sparse k-means algorithm will always flag 
a pre-determined number of observations as potential outliers.
The actual status of these suspicious cases is decided comparing
their weighted distance to the cluster centers with a 
threshold based on the distances to 
the cluster centers for all the observations.
Specifically, outliers are those
cases above the corresponding median plus 3.5 MADs of these
weighted distances. 
Using other reasonable values instead of 3.5 produced the
same conclusions. 
The resulting CERs are 0.06 for the robust 
sparse k-means with $\alpha = 10/78$, 
0.08 
when $\alpha = 5/78$ and when $\alpha = 1/78$, and 
0.10 for the non-robust sparse k-means ($\alpha = 0$). 
To simplify the comparison we restrict our attention to
SK-means and 
RSK-means with $\alpha = 10/78$. 

These sparse clustering methods can also be used to 
identify genes that play an important role for classifying
patients. 
In this regard, 62\% of the genes with positive weights
agree in both methods. On the other hand, 10\% of the
largest 
positive weights in each group (i.e. larger than the
corresponding
median non-zero weight) are different. 
The complete list genes selected by each method with
the corresponding weights is included in the suplementary
materials available on-line from our website 
\verb=http://www.stat.ubc.ca/~matias/pubs.html=. 

\begin{figure}[pt]
\begin{center}
\subfigure[SK-Means]{\resizebox{18cm}{3cm}{ \includegraphics{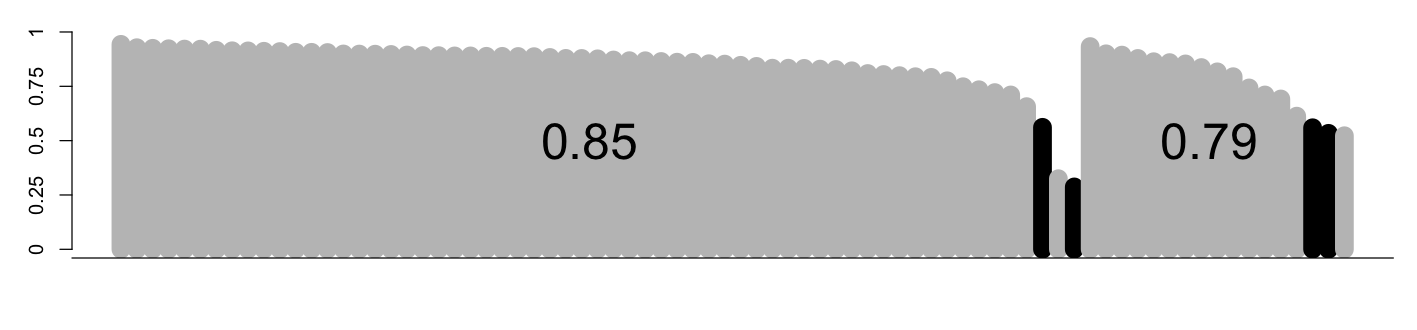}}}
\subfigure[RSK-means with $\alpha$=10/78]{\resizebox{18cm}{3cm}{ \includegraphics{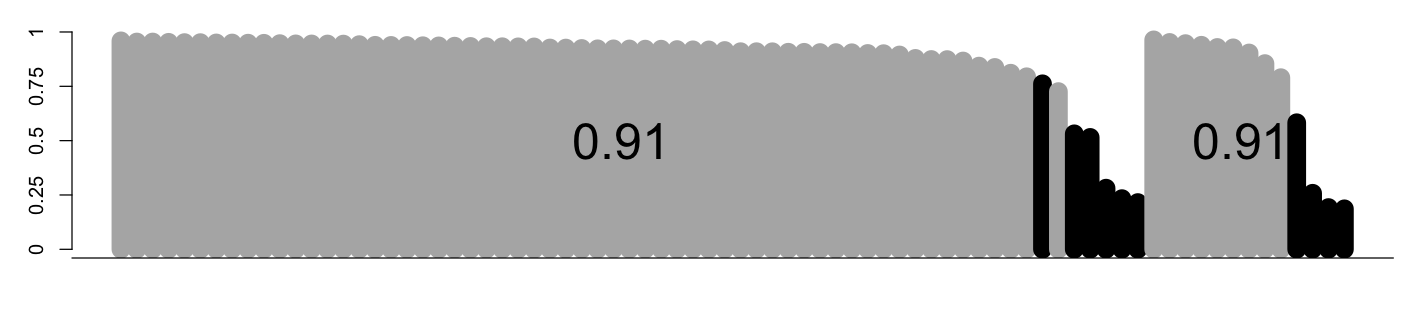}}}
\end{center}
\caption{
Silhouette plots for the partitions returned by SK-means and RSK-means with
$\alpha = 10/78$. 
Each bar corresponds to an observation and their heights correspond to 
their silhouette numbers. For each cluster, points are ordered according to their 
silhouette values. The numbers inside each cluster block indicate the 
corresponding 
average silhouette (over the observations in that cluster).
Higher values indicate stronger clusters. 
Darker bars indicate potentially outlying cases. 
}
\label{sil}
\end{figure}

To compare the strength of the clusters found by each of the two methods 
we use the following silhoutte-like index to measure how well separated
they are. 
For each observation $\bx_i$ let $a_i$ and $b_i$ be the squared distance 
to the closest and  
second
closest cluster centers, respectively. The ``silhouette'' of each point
is defined as $s_i = (b_i - a_i)/ b_i$. Note that $0 \le s_i \le 1$
with larger values indicating a stronger cluster assignment. 
The average silouhette of each cluster measures the overall strength 
of the corresponding cluster assignments. 
Figure \ref{sil} contains the silhouettes of the clusters
found by each method. Each bar corresponds to an observation and
with their height representing their silhouette. The numbers inside
each cluster denote the corresponding average silhouette for that
cluster. Darker lines identify outlying observations according to
the criterion mentioned above. 
Note that both clusters identified by RSK-means are considerably stronger than those
found by SK-means (the average silhouettes are 0.91 and 0.91, versus 0.85 and 0.79, respectively). 
Also note that the average silhouettes of 
the outliers identified by SK-means (0.44 and 0.55 for each cluster, respectively) 
are higher than those
identified by RSK-means (0.42 and 0.30 for each cluster, respectively). 
One can conclude that the outliers identified by RSK-means are more extreme 
(further on the periphery of their corresponding clusters) 
than those found by  
SK-means.

%

%

\section{Numerical results}
\label{simulations}

In this section we report the results of a simulation study to investigate
the properties of the proposed robust sparse K-means algorithm (RSK-means) and 
compare it 
with K-means (KM), trimmed K-means (TKM), and
sparse K-Means (SKM).
For this study we used our implementation
of RSK-means in the \verb=R= package
\verb=RSKC=, which is publicly 
available on-line at
\verb=http://cran.r-project.org=.

Our simulated datasets contain $n = 60$ observations generated from
multivariate normal distributions with covariance matrix equal to the identity.
We form three clusters of equal size by setting the mean vector to be $\bmu =
\mu \, \bb$, 
where 
$\bb \in \calR^{500}$ with 50 entries equal to 1 followed by 450 zeroes,
and $\mu = -1$, 0, and $1$, respectively.
Note that the clusters are determined by the first 50 features only (which we
call {\it clustering features}), the remaining 450 being {\it noise features}.

We will assess the performance of the different cluster procedures regarding
two outcomes: the identification of the true clusters and the identification of
the true clustering features.  To measure the degree of cluster agreement we use
the classification error rate (CER) proposed by \citet{CER}. 
Given two partitions of the data set (clusters) the CER is the proportion of
pairs of cases that are together in one partition and apart in the other.  To
assess the correct identification of clustering features we adapt the average
precision measure (reference?). This is computed as follows. First, sort the features
in decreasing order according the their weights and count the number of
true
clustering features appearing among the 50 higher-ranking features.


We consider the following two contamination configurations: 
\begin{itemize}
\item Model 1: We replace a single entry of a noise feature  ($\bx_{1, 500}$)
with an outlier at 25. 
\item Model 2: We replace a single entry of a clustering feature ($\bx_{1, 1}$)
with an outlier at 25. 
\end{itemize}
Several other configurations where we vary the number, value and location of
the outliers were also studied and are reported in the accompanying
supplementary materials. The general conclusions of all our studies agree with
those reported here. 

The $L_1$ bound for each algorithm was selected in such a way that 
approximately
50 features would receive positive weights when used on clean datasets. 
More specifically, we generated 50 data sets without outliers and, 
for each algorithm,  considered the following 11 
$L_1$ bounds: 5.7, 5.8, ..., 6.7. The one resulting in 
the number of non-zero weights closest to 50 was recorded.
In our simulation study we used the corresponding average of  
selected $L_1$ bounds for each algorithm. 
For both SK-means and RSK-means this procedure yielded an $L_1$ bound of
6.2. 
The proportion $\alpha$ of trimmed observations in TKM and RSK-means
was set equal to 1/60, the true proportion of outliers. 

\begin{figure}[htp]
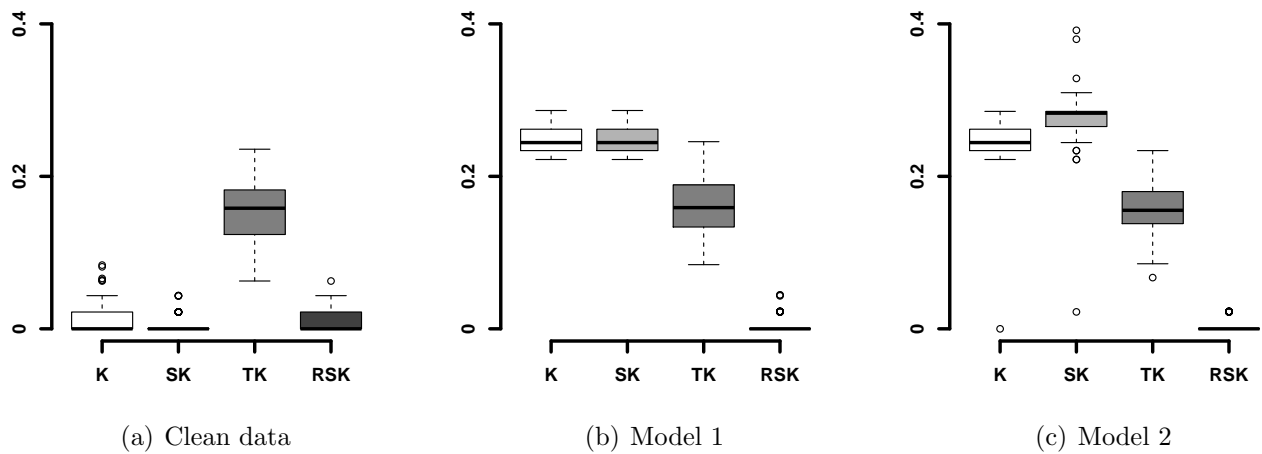

\begin{center}
\subfigure[Clean data]{\includegraphics[width=2.3in]{m0CERl150.png}}
\subfigure[Model 1]{\includegraphics[width=2.3in]{m1CERl150.png}}
\subfigure[Model 2]{\includegraphics[width=2.3in]{m2CERl150.png}}
\caption{
Boxplots of CERs calculated between the true partition and the
partitions from four algorithms. They correspond to 100 simulated
data sets of size $n=60$ and $p = 500$ features (with 50 true
clustering features). 
``K'', ``SK'', ``TK'' and ``RSK'' denote K-means, Sparse K-means, 
Trimmed K-means  and Robust Sparse K-means, respectively.
}
\label{cers}
\end{center}
\end{figure}

\begin{figure}[htp]
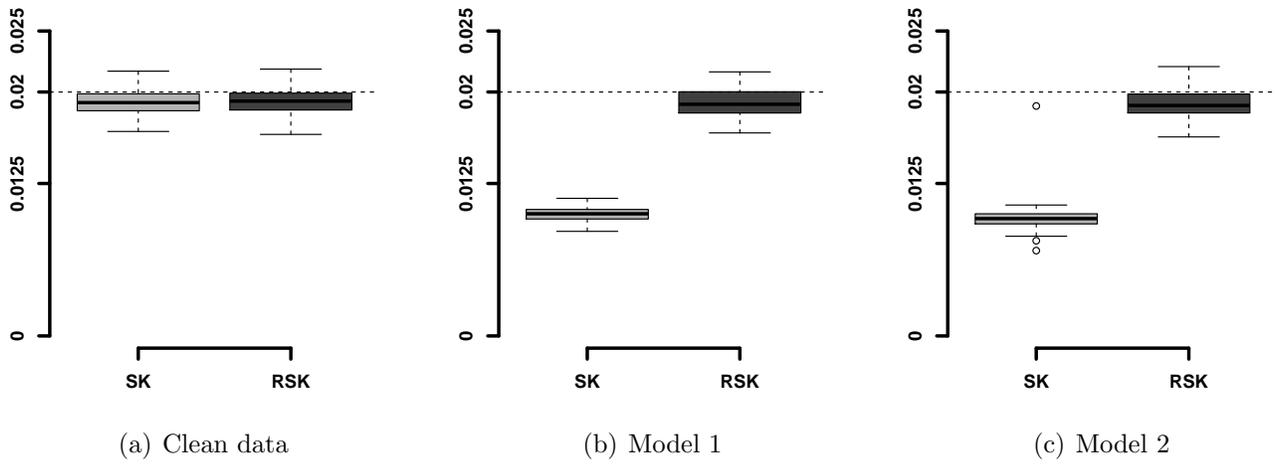

\begin{center}
\subfigure[Clean data]{\includegraphics[width=2.3in]{m0wl150.png}}
\subfigure[Model 1]{\includegraphics[width=2.3in]{m1wl150.png}}
\subfigure[Model 2]{\includegraphics[width=2.3in]{m2wl150.png}}
\end{center}
\caption{
The boxplots of median proportions of weights on the 50 clustering features over 100 simulations. 
The dotted line represents the ideal amount of weights on each clustering feature.
K=K-means, SK=SK-means,TK=TK-means, RSK=RSK-means.}
\label{weight}
\end{figure}

We generated 100 datasets from each model. 
Figure \ref{cers} shows the boxplots of CERs between the true
partition and the partitions returned by the algorithms.
When the data do not contain any outliers we see that 
the performance of K-means, SK-means, and RSK-means
are comparable. 
The results of Models 1 and 2 show that the 
outliers affect the peformance of K-means and SK-means, and also
that the presence of 450 noise features upsets the performance
of the TK-means. However, the RSK-means algorithm retains small values of
CER for both types of contamination. 
To compare the performance of the different algorithms with 
regards to the features selected by them we consider the 
median weight assigned to the true clustering features. 
The results are shown in Figure \ref{weight}. We can see
that when there are no outliers in the data both the SK-means
and RSK-means algorithms assign very similar weights to the
correct clustering features. The presence of a single outlier,
however, results in the SK-means algorithm to assign much
smaller weights to the clustering features. 
%
%
Table 
\ref{weight-table} below contains 
the average number of non-zero weights returned
by each algorithm, and
average number of true clustering
features among the 50 features receiving highest weights. 
When the data are clean, both SK-means and RSK-means 
return approximately 50 clustering features, and they are
among the ones with highest weights. A single outlier
(with either contamination Model 1 or 2) results in 
SK-means selecting almost all 500 features, while 
RSK-means remains unaffected.
\begin{table}[hptb]
\begin{center}
\begin{tabular}{cc|rr} \hline \hline
 & &\multicolumn{1}{|c}{Non-zero weights}&\multicolumn{1}{c}{Average precision}\\ \hline \hline
No outliers & RSK-means & 49.2 (1.28)   & 48.9 (0.31) \\ 
        & SK-means  & 49.3 (1.12)   & 48.9 (0.26) \\ \hline
Model 1 & RSK-means & 49.2 (1.00)   & 48.9 (0.31) \\ 
        & SK-means  & 498.5 (1.18)  & 47.5 (0.64) \\ \hline
Model 2 & RSK-means & 49.2 (1.11)   & 49.0 (0.17) \\ 
        & SK-means  & 494.0 (44.0) & 47.8 (1.00) \\ 
\hline \hline
\end{tabular}
\caption{The first column contains the average number of non-zero
  weights over the 100 data sets (SD in parentheses). The
  second column shows the
average number of true clustering features among the 
50 features with largest weights (SD in parentheses).}
\label{weight-table}
\end{center}
\end{table}

\section{Conclusion}

In this paper we propose a robust algorithm to simultaneously identify clusters
and features using K-means. The main idea is to adapt the Sparse K-means
algorithm of \citet{Dan} by trimming a fixed proportion of observations that
are farthest away from their cluster centers (using the approach of the Trimmed
K-means algorithm of \citet{trimmed}).  Sparcity is obtained by assigninig
weights to features and imposing an upper bound on the $L_1$ norm of the vector
of weights. Only those features for which the optimal weights are positive are
used  to determine the clusters.  Because possible outliers may contain
atypical entries in features that are being downweighted, our algorithm also
considers the distances from each point to their cluster centers using all
available features.  Our simulation studies show that the performance of our
algorithm is very similar to that of Sparse K-means when there are no oultiers
in the data, and, at the same time, it is not severely affected by the presence
of outliers in the data. We used our algorithm to identify relevant clusters in
a data set containing gene expressions for breast cancer patients, and we were
able to find interesting biological clusters using a very small proportion of
genes. Furthermore, the clusters found by the robust algorithm are stronger
than those found by Sparse K-means.  Our robust Sparse  K-means algorithm is
implemented in the \verb=R= package \verb=RSKC=, which is available at the
\verb=CRAN= repository.

\bibliographystyle{abbrvnat}
\bibliography{BIB}






\end{document}